\documentclass{article}

\usepackage{microtype}
\usepackage{graphicx}
\usepackage{subfigure}
\usepackage{booktabs} 

\usepackage{hyperref}

\usepackage[accepted]{icml2025}

\usepackage{amsmath}
\usepackage{amssymb}
\usepackage{mathtools}
\usepackage{amsthm}
\usepackage[T1]{fontenc}
\usepackage[utf8]{inputenc}
\usepackage[capitalize,noabbrev]{cleveref}

\theoremstyle{plain}
\newtheorem{theorem}{Theorem}[section]

\theoremstyle{definition}
\newtheorem{definition}[theorem]{Definition}

\theoremstyle{remark}

\usepackage[textsize=tiny]{todonotes}

\usepackage{natbib}
\icmltitlerunning{}

\begin{document}
\twocolumn[
\icmltitle{From Robotics to Sepsis Treatment: Offline RL via Geometric Pessimism}



\icmlsetsymbol{equal}{*}

\begin{icmlauthorlist}
\icmlauthor{Sarthak Wanjari}{}
\end{icmlauthorlist}



\icmlkeywords{Machine Learning, ICML}

\vskip 0.3in
]





\begin{abstract}
Offline Reinforcement Learning (RL) promises the recovery of optimal policies from static datasets, yet it remains susceptible to the overestimation of out-of-distribution (OOD) actions, particularly in fractured and sparse data manifolds. Current solutions necessitate a trade-off between computational efficiency and performance. Methods like CQL offer rigorous conservatism but require tremendous compute power while efficient expectile-based methods like IQL often fail to correct OOD errors on pathological datasets, collapsing to Behavioural Cloning. In this work, we propose Geometric Pessimism, a modular, compute-efficient framework that augments standard IQL with density-based penalty derived from $k$-nearest-neighbour distances in the state-action embedding space. By pre-computing the penalties applied to each state-action pair, our method injects OOD conservatism via reward shaping with a $\mathcal{O}(1)$ training overhead to the training loop.

Evaluated on the D4RL MuJoCo benchmark, our method, Geo-IQL outperforms standard IQL on sensitive and unstable \texttt{medium-replay} tasks by over 18 points, while reducing inter-seed standard-deviation by 4$\times$. Furthermore, Geo-IQL does not degrade performance on stable manifolds. Crucially, we validate our algorithm on the MIMIC-III Sepsis critical care dataset. While standard IQL collapses to behaviour cloning, Geo-IQL demonstrates active policy improvement. Maintaining safety constraints, it achieves 86.4\% terminal agreement with clinicians compared to IQL's 75\%. Our results suggest that geometric pessimism provides the necessary regularisation to safely overcome local optima in critical, real-world decision systems.
\end{abstract}
\section{Introduction}
\label{sec:intro}
Reinforcement Learning (RL) has achieved superhuman performance in simulated domains, solving complex strategy games such as Go and Dota 2 \cite{berner2019dota, silver2017mastering}. In these environments, RL agents learn through an “online” paradigm, i.e. learning occurs through continuous interaction with the environment, where data is generated sequentially as a result of the agent's own actions and is then used to update the policy \cite{murphy2024reinforcement}. If an agent crashes a virtual car or loses a game then the simulation is simply reset. However, this reliance on trial-and-error has rendered RL effectively useless in high-stakes, real-world domains such as robotic surgery, sepsis management and autonomous driving \cite{levine2020offline}. In these domains a single exploratory action no longer leads to a trivial “Game Over” screen, but instead may cause catastrophic hardware failure and even patient mortality. To enable safe deployment in such high-stakes settings, RL must fundamentally adapt beyond its trial-and-error paradigm.

To address this safety gap a new paradigm known as Offline RL has emerged. Instead of iteratively interacting with the world in a dynamic setting, Offline RL algorithms learn policies entirely from static, historical datasets \cite{levine2020offline}, such as previously collected hospital logs (e.g. MIMIC-III Critical Care dataset) or human driving data for the purpose of autonomous driving.

However, Offline RL introduces a subtle but dangerous failure mode, the problem of Distributional Shift \cite{levine2020offline}, that results in value overestimation which can have dire consequences. In classical RL, algorithms assume that optimistic estimates for untried actions are acceptable because the agent can directly attempt them by interacting with the environment and receive its respective reward. If the reward is low, the algorithm learns to reduce the probability of attempting that action.

On the contrary, in a closed dataset setting, this learning process breaks down because the agent cannot interact with the environment to correct its mistakes during training. When the policy attempts an action outside its training distribution (out-of-distribution, OOD), it has the potential to hallucinate and assign arbitrarily high rewards simply because no data exists to suggest otherwise \cite{kumar2019stabilizing}. Since the agent cannot gather new information, these errors persist and propagate via the Bellman backup, which is the foundational core of RL, leading to permanent overestimation of poor actions. Consequently, the agent assumes that unknown actions are the best actions (due to its erroneously high estimated value) causing it to confidently execute dangerous, erratic behaviours in the real world upon deployment \cite{kumar2019stabilizing}.

\section{Background}
The landscape of offline RL is generally categorised into two camps: Constrain-Based Methods that restrict the policy to the dataset, and Pessimism-based methods that alter the value function to penalise actions from unknown regions. 

\textbf{Conservative Q-Learning (CQL):} Representing the Pessimistic camp, CQL is currently the gold standard for safety due to its conservative actions and is canonically a very strong baseline when researching new methods. It operates by adding a regularisation term to the objective function that actively “pushes down” the Q-Values of OOD actions \cite{kumar2020conservative}. Although effective at preventing overestimation, this comes at a very high computational cost as CQL requires sampling unseen actions during every training step to compute the log-sum-exp regulariser, resulting in slower convergence as well \cite{kumar2020conservative}. 

\textbf{Implicit Q-Learning (IQL):} To address the computational inefficiency of CQL, IQL avoids querying OOD actions altogether \cite{kostrikov2021offline}. Instead it treats the value function as a regression problem to estimate the Q value. It estimates the optimal Q-value functions via expectile regression, allowing IQL to remain entirely “in-sample” offering significant speed advantages and stability \cite{kostrikov2021offline}. However, IQL lacks an explicit mechanism to detect OOD data. As a result in “fractured” dataset where gaps in the state space coverage exist, IQL could struggle to estimate the optimal Q-Values, leading to sub-par performance.

\textbf{Geometric Uncertainty:} While some prior works have explored uncertainty estimation, they typically rely on ensembles of neural networks and use the variance as a proxy for risk \cite{clements2019estimating}. These ensembles are computationally very expensive to train and run. Our work diverges from this by utilising geometric distance between data points in the embedding space as a proxy instead and using this as a form of penalisation. This approach is deterministic, explainable and as we demonstrate can be pre-computed to maintain the runtime efficiency of IQL while recovering the safety guarantees of CQL.

\section{Hypothesis}
We hypothesise that geometric distance in the joint state-action embedding space can be used a proxy for the epistemic uncertainty in Offline Reinforcement Learning. Specifically, we predict that by penalising the Q-value estimates of actions based on their Euclidean distance to the nearest $k$ Neighbors in the training set, we can reduce the overestimation error and hallucination without requiring the computationally expensive sampling procedures used in current state-of-the-art methods such as CQL.

\section{Methodology}
We propose Geo-IQL (Geometric Implicit Q-Learning), an algorithm designed to inject geometric safety constraints into the already efficient IQL algorithm. The algorithm operates in a four-stage pipeline: (1) Embedding, (2) Geometric Uncertainty Estimation, (3) Robust Standardisation, and (4) Reward Penalisation.

\subsection{State-Action Embedding}
To measure how “familiar” a state-action pair is (i.e. taking a specific action in a given state), we first map the (state, action) data into a joint embedding space. We define the embedding vector $\phi(s,a)$ by concatenating the Gaussian-normalised state vector with the raw action vector:
\begin{equation}
\phi(s,a) = [\text{norm}(s), a] \in \mathbb{R}^{d_s + d_a}
\end{equation}
\textbf{Design Decision:} We explicitly excluded action normalisation from this embedding. In control tasks, the magnitude of the action (e.g. how hard a motor spins) carries physical significance. Preserving the raw action scale ensures that Euclidean distance in this space reflects true physical divergence in control inputs.

\begin{figure}[ht]
\begin{center}
\centerline{\includegraphics[width=\columnwidth]{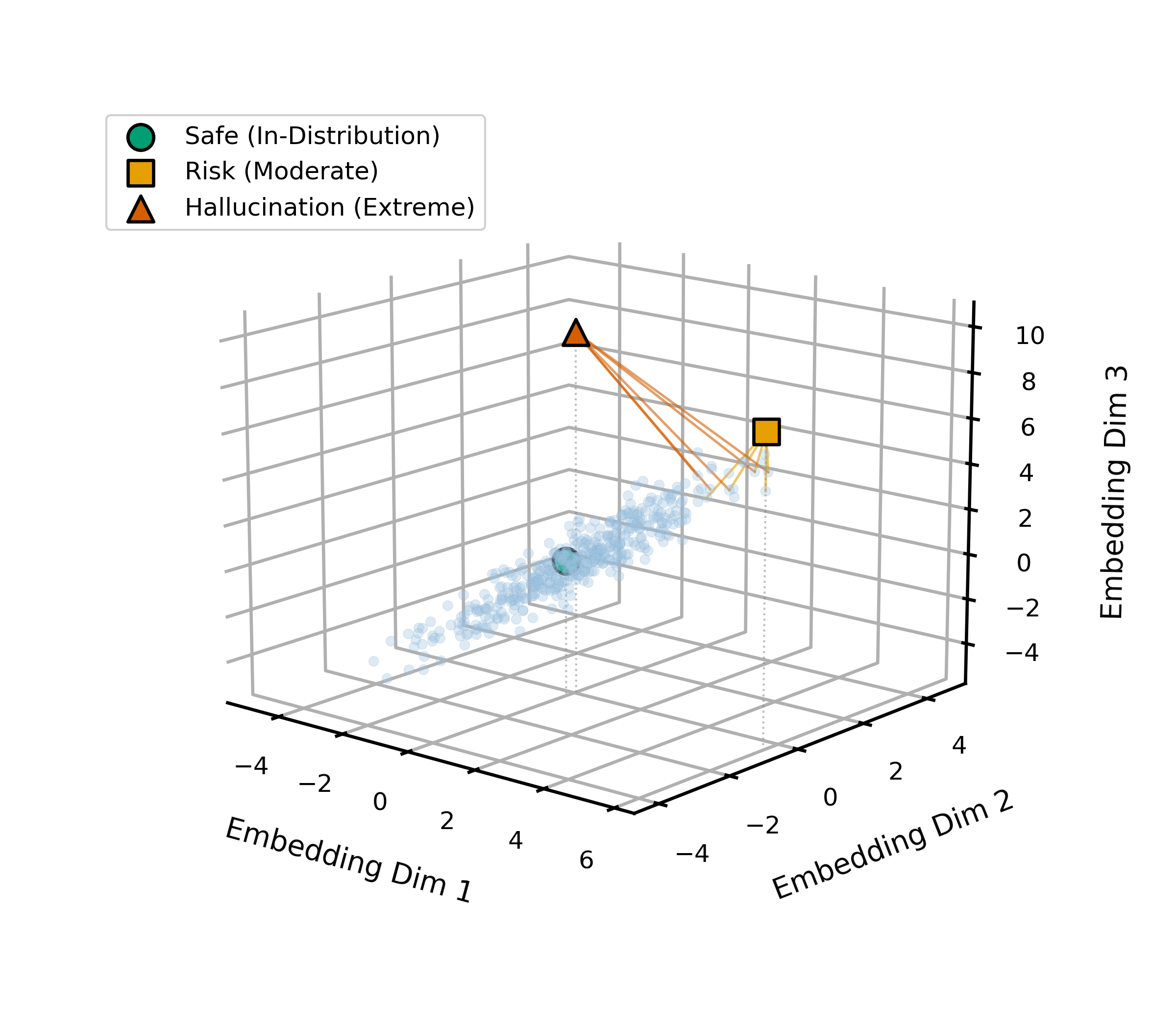}}
\vskip -0.3in
\caption{\textbf{3-Dimensional} visualisation of using geometry as a proxy for epistemic uncertainty.}
\label{fig:adaptive_geometry}
\end{center}
\vskip -0.2in
\end{figure}

\subsection{Geometric Uncertainty}
We hypothesise that epistemic uncertainty, the risk, is proportional to the distance from the training data manifold. A data point far from the dataset manifold is OOD and one that is in a dense region is in distribution. For any query pair $(s,a)$, we calculate a raw uncertainty score $\tilde{U}(s,a)$, the mean distance to the $k$ closest $(s,a)$ points using a k-Nearest Neighbors (KNN) approach. Let $\mathcal{N}_k$ be the set of the $k$ closest data points in the static dataset $\mathcal{D}$:
\begin{equation}
\tilde{U}(s,a) = \frac{1}{k} \sum_{j \in \mathcal{N}_k} || \phi(s,a) - \phi_j ||_2
\end{equation}
We selected $k=10$ to balance local sensitivity with robustness against individual noise points.

\subsection{Robust Standardisation}
Raw Euclidean distances vary wildly depending on the environment dimensions. To make our algorithm transferable across different tasks, we must standardise these scores. Standard mean-variance normalisation is susceptible to outliers. Therefore, we engineer a robust standardisation utilising Median Absolute Deviation (MAD).

We define a “safe threshold”, $\tau$, as the $\alpha$-quantile of distances in the dataset ($\alpha=0.3$ based on preliminary tuning). This explicitly categorises the closest 30\% of the data as the “Safe Core”, receiving a penalty of 0. This then implies that the remaining 70\% of the dataset is subject to penalisation. However the our standardisation ensures this is not a binary punishment, but a graded risk surface. We compute the robust spread $\sigma_{MAD}$ with a small $\epsilon$ to ensure division by zero does not occur:
\begin{equation}
\sigma_{\text{MAD}} = \text{median}(\{|\tilde{U}_i - \tau |\}_{i=1}^N) + \epsilon
\end{equation}
The standardised uncertainty score $U(s,a)$ is then:
\begin{equation}
U(s,a) = \frac{\tilde{U}(s,a) - \tau}{\sigma_{\text{MAD}}}
\end{equation}
This step ensures that $U(s,a)$ represents a statistical z-score that is robust to outliers, effectively filtering out noise.

\subsection{Density-Adaptive Penalisation}
Finally, we incorporate this signal into the learning objective. We aim for a penalty that is negligible in safe regions but grows aggressively in unknown regions. We implement a Density-Adaptive Scaling factor $\rho(s,a)$:
\begin{equation}
\rho(s,a) = \frac{1}{1 + \max(0, U(s,a))}
\end{equation}
This factor decays from 1 to 0 as uncertainty increases. We use this to modulate the penalty weight $\lambda_{adapt}$ from $\lambda_{base}$ a hyper-parameter:
\begin{equation}
\lambda_{\text{adapt}} = \lambda_{\text{base}} \cdot (2 - 1.5 \rho(s,a))
\end{equation}
\begin{equation}
    \lambda_{adapt} \in [0.5 \lambda_{base}, 2\lambda_{base}]
\end{equation}

The final reward function used to train the agent is:
\begin{equation}
r_{\text{geo}}(s,a) = r(s,a) - \lambda_{\text{adapt}} \cdot \max(0, U(s,a))
\end{equation}
\textbf{Interpretation:} When the agent operates within a dense data region, ($U\approx0$), the penalty is zero. As the agent drifts into OOD regions ($U>0$), the penalty activates and scales up, reducing the perceived value of that action in the Bellman backup. If the OOD score is low we penalise lightly, if it is high we penalise heavily. This process allows us to have a unique penalty for each data point, increasing rigour and accuracy. This mathematically forces the agent to return to the safety of the data manifold. \textit{(See Fig 2.)}

\begin{figure}[ht]
\vskip 0.2in
\begin{center}
\centerline{\includegraphics[width=\columnwidth]{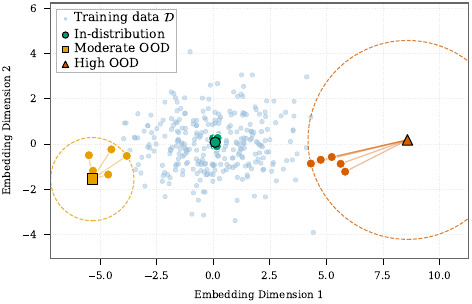}}
\caption{\textbf{Visualising the Adaptive Safety Mechanism.} The blue cloud represents the training data manifold. 
\textbf{Green Point:} A query inside the dense region. The algorithm detects near-zero distance to neighbours and applies no penalty.
\textbf{Yellow Point:} A query slightly off the manifold. This triggers the adaptive lambda, applying a mild correctional penalty.
\textbf{Red Point:} A query far from the data. Effectively, OOD. The algorithm measures a large mean Euclidean distance from its neighbours and applies a heavy penalty to reject this action entirely.}
\label{fig:adaptive_geometry}
\end{center}
\vskip -0.25in
\end{figure}

\subsection{Computational Efficiency}
Unlike previous methods that calculate constraints in the training loop, Geo-IQL allows for pre-computation. Since the dataset $\mathcal{D}$ is static, we compute the penalties for all transition tuples and store them in a lookup table \textit{before} the neural network training even begins.

Consequently, during the training loop, looking up the penalties is an $\mathcal{O}(1)$ operation as the penalties are stored on a tensor. This allows Geo-IQL to train at the same speed as the baseline IQL, ensuring efficiency and scalability

\subsection{Algorithmic Formulation}
\subsubsection{IQL: Target and Loss}
IQL utilises Expectile Regression to learn the value function without querying out-of-sample actions. The framework consists of Twin Q-Networks $Q_{\theta_{1,2}}$, a Value network $V_{\psi}$, and an Actor $\pi_{\phi}$. The standard critic objective minimises the Mean Squared Error (MSE) against a target constructed from the value network:
\begin{equation*}
    y = r(s,a) + \gamma V_{\psi}(s')
\end{equation*}
\begin{equation*}
    \mathcal{L} = \mathbb{E}_{(s,a,r,s') \sim \mathcal{D}} [ (Q_{\theta}(s,a) - y)^2 ]  
\end{equation*}

\subsubsection{Geo-IQL: Modifications}
We introduce a geometric penalty term to the reward function. Let $r_{geo}(s,a) = r(s,a) - \lambda_{\text{adapt}} \cdot \max(0, U(s,a))$ be penalised reward. The Geo-IQL critic target becomes:
\begin{equation*}
    y_{geo} = r_{geo} + \gamma V_{\Phi}(s')
\end{equation*}

Crucially, the Value ($V_{\psi}$) and Actor ($\pi_{\phi}$) update rules remain identical to the original IQL formulation. This ensures that the geometric module acts as a \textit{strictly} additive constraint on the value estimation, preserving the stability of the underlying algorithm. The entire algorithm is presented in \textit{Algorithm 1}.

\begin{algorithm}[tb]
   \caption{Geo-IQL with Adaptive Geometric Pessimism}
   \label{alg:geo_iql_adaptive}
\begin{algorithmic}
   \STATE {\bfseries Input:} Dataset $\mathcal{D}$, Neighbors $k$, Base scale $\lambda_{\text{base}}$,
   Threshold percentile $\alpha$
   
   \STATE \COMMENT{\textcolor{blue}{\textbf{Phase 1: Penalty Pre-Computation}}}
   \STATE Embed state-action pairs: $Z \leftarrow \{\phi(s, a) \mid (s, a) \in \mathcal{D}\}$
   \STATE Build $k$-Nearest Neighbour Index on $Z$
   
   \STATE \textbf{Compute Geometry:}
   \STATE \quad Compute raw scores $\tilde{U} \leftarrow \text{Mean-kNN-Distance}(Z)$
   \STATE \quad Compute statistics: $\tau \leftarrow \text{Quantile}_\alpha (\tilde{U})$, $\sigma \leftarrow \text{MAD}(\tilde{U})$
   \STATE \quad Normalise: $U \leftarrow (\tilde{U} - \tau) / \sigma$
   
   \STATE \textbf{Adaptive Scaling:}
   \STATE \quad Estimate local density $\rho$ for each point in $Z$
   \STATE \quad Compute adaptive scale: $\lambda_{\text{adapt}} \leftarrow \lambda_{\text{base}} \cdot (2 - 1.5 \rho)$
   \STATE \quad Store final penalty: $P \leftarrow \lambda_{\text{adapt}} \cdot \text{max}(0,U)$ in a Lookup Table
   
   \STATE \COMMENT{\textcolor{blue}{\textbf{Phase 2: Offline Training Loop}}}
   \REPEAT
   \STATE Sample batch $\{(s, a, r, s')\}$ and indices $I$ from $\mathcal{D}$
   
   \STATE \textbf{Pessimistic Reward:}
   \STATE \quad Retrieve precomputed penalties: $p \leftarrow P[I]$
   \STATE \quad Apply penalty: $r' \leftarrow r - p$
   
   \STATE \textbf{Update:}
   \STATE \quad Perform standard IQL updates using modified reward $r'$
   \UNTIL{Convergence}
\end{algorithmic}
\end{algorithm}

\subsection{Theoretical Justification}
The geometric penalty is also theoretically grounded in Lipschitz continuity.
The key insight is that Lipschitz smooth functions cannot change arbitrarily fast. If our learned $\widehat{Q}$ is accurate on the training data, its error at a new point $(s,a)$ is bounded by how far that point is from the data. By subtracting a penalty proportional to this distance, we provably obtain a lower bound on the true value. We guide the reader to the formal proof and intuitive visualisation in the appendix.

\section{Experimental Setup}
To validate our hypothesis we conduct a comparative analysis across two distinct domains: high dimensional robotic control and critical healthcare management. These domains have a continuous and discrete action space respectively.

\subsection{Robotic Locomotion (D4RL)}
\textbf{Environment:} We evaluate on the D4RL MuJoCo suite \cite{fu2020d4rl}, the standard benchmark for continuous control in offline RL. Specifically, we focus on Hopper, Walker2d, and HalfCheetah (see Figures 3, 4 and 5) \cite{park2024tackling} which require controlling articulated robots to achieve stable locomotion. The tasks are as follows:
\begin{enumerate}
    \item \textbf{HalfCheetah:} A 2-Dimensional bipedal robot with the task of running forward as fast as possible
    \item \textbf{Hopper:} A one-legged robot that must hop forward without falling over.
    \item \textbf{Walker2d:} A two-legged robot that walks forward
\end{enumerate}

Each of these robots are controlled by a policy recovered from the static dataset by the algorithm. Performance over these locomotion environments are reported as a Normalised Score, where 0 corresponds to a random policy and 100 to an expert policy. Scores exceeding 100 indicate the agent has outperformed the data-collecting policy that the dataset was generated from.
\begin{figure}[t]
\centering
\begin{subfigure}{}
    \includegraphics[width=0.5\columnwidth]{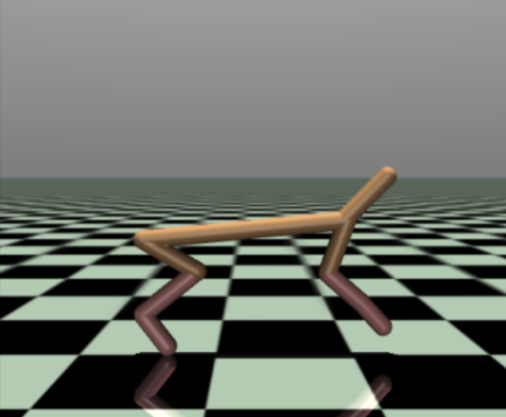}
    \vskip -0.15in
    \caption{HalfCheethah}
\end{subfigure}
\vskip 0.1in
\begin{subfigure}{}
    \includegraphics[width=0.5\columnwidth]{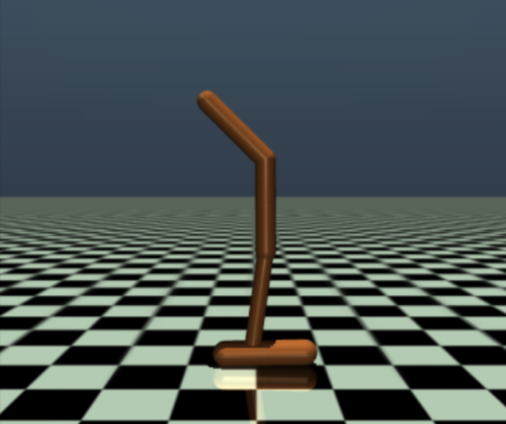}
    \vskip -0.15in
    \caption{Hopper}
\end{subfigure}
\vskip 0.1in
\begin{subfigure}{}
    \includegraphics[width=0.5\columnwidth]{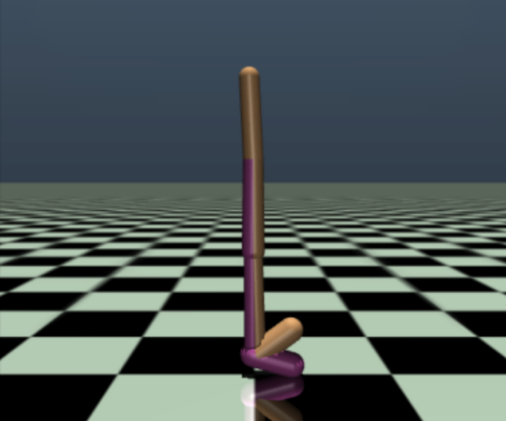}
    \vskip -0.15in
    \caption{Walker2d}
\end{subfigure}
\end{figure}

We utilise the medium-replay datasets for all tasks. These datasets consist of samples in the replay buffer observed during training until the policy reaches the “medium” level of performance from scratch \cite{fu2020d4rl}. Consequently, the data is highly fractured with mixed quality transitions, presenting a significant challenge for algorithms that rely on “in sample” learning such as IQL. Making it the perfect choice to test the performance of our algorithm to recover the good actions while ignoring the bad ones. 

\subsection{Sepsis Management (MIMIC-III)}
\textbf{Environment:} Additionally, to assess performance in a high-stakes, real-world setting we utilise the MIMIC-III Sepsis Cohort \cite{PhysioNet, alistair_johnson_2025_15272720, johnson2018mimic, Johnson2023-sj}. The objective is to prescribe optimal dosages of vasopressors and intravenous fluids (discrete action space $|\mathcal{A}| = 25$) to maximise patient survival. Since online evaluation is impossible, as we do not test on live patients. We report metrics across three categories: Clinician Agreement, Policy Quality and Safety.
\subsubsection{Clinician Agreement}
\begin{enumerate}
    \item \textbf{Policy-Clinician Agreement:} How often does the policy agree the action taken by the clinician for a given health state.
    \item \textbf{Probability of Clinician Action:} At what probability would the policy take the action that the clinician took? This essentially represents the policy's belief on how reasonable the clinician's actions were. High probability signifies that the action were very reasonable and a low probability means that the action were highly suboptimal from the perspective of the policy
    \item \textbf{KL-Divergence:} Measuring how different is the probability distribution of action form the clinician's in a given state.
\end{enumerate}

\subsubsection{Policy Quality}
\begin{enumerate}
    \item \textbf{Q-Improvement:} The difference between the Q-Values of policy's action and the clinicians.
    \begin{equation*}
        \Delta Q = \mathbb{E} [Q(s, \pi_{policy}) - Q(s, a_{clinician})]
    \end{equation*}
    When positive and rising it signifies that on average the algorithm makes better decisions than the clinician.

    \item \textbf{Policy Entropy: } A confidence metric of the policy. Lower the entropy, more concentrated the probability is over a certain action. The algorithm is more confident in its decision.
    \begin{equation*}
        \mathcal{H}(\pi(\cdot | s)) = -\sum_{a \in \mathcal{A}} \pi(a|s)\log(\pi(a|s))
    \end{equation*}

    \item \textbf{Terminal State Agreement: } This measures how often the policy matched the clinician's actions in the last few steps. In Sepsis treatment, the “middle” of a trajectory is often noisy as there are many ways to keep a stable patient stable. However, the end is critical. If the patient survived, then clinician’s final actions likely stabilised the patient enough for discharge. A high agreement here signifies that the policy learned “how to successfully discharge a patient”. 
    
    On the other hand, if the patient died, the clinician was likely responding to acute clinical deterioration. High agreement here signifies that the policy successfully recognised a medical crisis and attempted actions that a clinician would deem the best. 

\end{enumerate}
\subsubsection{Safety}
\begin{enumerate}
    \item \textbf{Dose Deviation: } This metric quantifies the deviation of the doses prescribed by the policy from the clinician true action.
    \item \textbf{Extreme action agreement: } The degree of similarity that the policy has for extreme action such as no treatment at all of very high doses of vasopressors and fluids at once.
\end{enumerate}

\begin{table*}[t]
\centering
\small
\setlength{\tabcolsep}{6pt}
\caption{D4RL MuJoCo locomotion, Normalised Scores. 
Mean $\pm$ standard deviation over 5 seeds.}
\label{tab:d4rl}
\begin{tabular}{lcccc}
\toprule
Task & BC & CQL & IQL & Geo-IQL \\
\midrule
halfcheetah-medium-replay-v2  
& 27.69 $\pm$ 10.92 
& \textbf{45.41 $\pm$ 0.81}
& 43.68 $\pm$ 4.15  
& 42.52 $\pm$ 3.04 \\

hopper-medium-replay-v2  
& 51.87 $\pm$ 20.26 
& 82.60 $\pm$ 21.10
& 80.09 $\pm$ 21.80 
& \textbf{98.94 $\pm$ 5.33} \\

walker2d-medium-replay-v2 
&  43.17 $\pm$ 25.77
& 78.28 $\pm$ 18.85 
& 80.17 $\pm$ 17.89 
& \textbf{82.10 $\pm$ 13.39} \\
\bottomrule
\end{tabular}
\end{table*}


\begin{table*}[t]
\centering
\caption{Offline Evaluation on MIMIC-III Sepsis}
\label{tab:mimic-all}
\begin{tabular}{@{}lccccccc@{}}
\toprule
& \multicolumn{3}{c}{Clinician Agreement} & \multicolumn{3}{c}{Policy Quality}\\
\cmidrule(lr){2-4} \cmidrule(lr){5-7}
Algorithm & P-C Agr. & P(Clin. Act.) & KL-Div. & $\Delta Q$ & Entropy & Term. Agr.\\
\midrule
IQL     & 68.68\% & 0.5741 & 0.0057 & -0.0169 & 0.8255 & 75.02\%\\
Geo-IQL & 64.16\% & 0.5679 & 0.0052 & \textbf{0.0138} & \textbf{0.6924} & \textbf{86.39}\%\\
\bottomrule
\end{tabular}
\end{table*}

\begin{table}[t]
\centering
\caption{Safety Metrics}
\label{tab:safety}
\small
\begin{tabular}{@{}lccc@{}}
\toprule
Algorithm & Vaso. Dev. & Fluids Dev. & Ext. Agr. \\
\midrule
IQL     & 0.496 & 0.02 & 98.42\% \\
Geo-IQL & 0.572 & 0.046 & 97.86\% \\
\bottomrule
\end{tabular}
\end{table}

\subsection{Implementation and Hardware Constraints}
All algorithms were implemented using the CORL (Clean Offline RL) library \cite{tarasov2023corl, tarasov2022corl}, to ensure reproducibility. Link to codebase can be found in the appendix. All experiments were conducted on a single consumer-grade laptop GPU, NVIDIA RTX 5060. This constraint demonstrates that Geo-IQL's $\mathcal{O}(1)$ access to pre-computed penalty mechanism makes safe AI accessible without requiring industrial-scale compute clusters (unlike ensemble methods). Additionally we report the mean and standard deviation over 5 random seeds to ensure statistical significance.
\section{Results and Analysis}
\subsection{D4RL MuJoCo}
As shown in \textit{Table 1}, Geo-IQL achieves substantial gains in the fractured \texttt{hopper-medium-replay-v2} domain, outperforming standard IQL by \textbf{over 18} points while reducing inter-seed standard deviation by approximately $\mathbf{4 \times}$. Notably, Geo-IQL also surpasses the conservative baseline CQL, yet maintains the computational efficiency of IQL, avoiding the expensive iterative regularisation that is inherent to CQL.

The learning curves in \textit{Fig. 6} further highlight the improved sample efficiency of our method. We see that Geo-IQL converges to high-performing policies in significantly fewer gradient steps than both baselines. We also evaluate Behaviour Cloning (BC), essentially supervised learning on the static dataset to establish a strict bottom line performance that any offline RL algorithm should surpass.

In the more stable multi-legged locomotion tasks (\texttt{halfcheetah}, \texttt{walker2d}), Geo-IQL achieves performance statistically equivalent to IQL, with overlapping standard deviation. This behavior is expected and desirable. It confirms that our geometric penalty is adaptive, applying necessary conservatism in sparse, unstable regimes, such as Hopper's hyper-sensitive survival dynamics where the smallest errors and overestimation leads to early termination, all while not degrading performance in the denser, stable manifolds of HalfCheetah.

The slight mean differences observed are well within the inherent variance of the \texttt{medium-replay} datasets, which contain heterogeneous mixtures of noisy exploration and competent trajectories. This is strong evidence that Geo-IQL acts as a robust augmentation, strictly improving stability in pathological settings without compromising capability in standard domains.

\begin{figure}[t]
    \centering
    \includegraphics[]{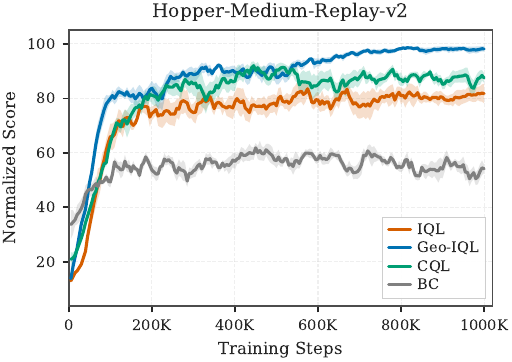}
    \vskip -0.1in
    \caption{Performance over 1M training steps.}
    \label{fig:myfigure}
\end{figure}

\begin{figure}[t]
\begin{center}
\centerline{\includegraphics[width=\columnwidth]{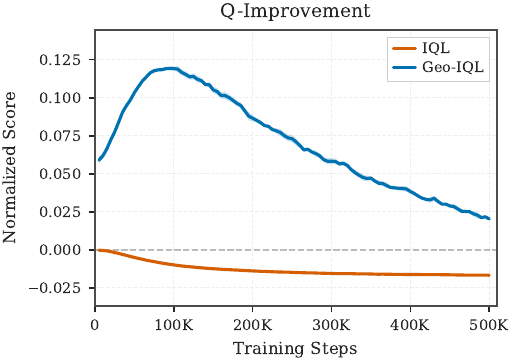}}
\caption{\textbf{Q-Improvement} of IQl and Geo-IQL}
\label{fig:adaptive_geometry}
\end{center}
\vskip -0.2in
\end{figure}

\subsection{MIMIC-III Sepsis}
Standard IQL achieves higher overall agreement (68.7\% vs 64.2\%), suggesting it is performing primarily as a high-fidelity mimic of the average clinical behaviour. However, this passivity comes at a cost. IQL fails to improve upon the clinician's value estimate as demonstrated by the negative $\Delta Q$ metric, a value of -0.0169 and exhibits higher uncertainty with Entropy being 0.83.

In contrast, Geo-IQL demonstrates targeted optimisation. While it agrees less often on intermediate steps, reflected by its active adjustment of vasopressor dosages to maximise return. It achieves a positive value improvement, $\Delta Q = +0.0138$. Furthermore, from \textit{Fig 7.} we see that Q-Improvement for Geo-IQL increases and tapers of (as its estimate of clinician slowly catches up to the policy's estimate of itself) but is still positive. On the other hand, IQL is always negative. 

Crucially, this optimisation does not come at the cost of safety. Geo-IQL achieves a significantly higher Terminal State Agreement (86.39\% vs 75.02\%), indicating that while it may optimise the treatment path, it aligns far more robustly with clinician decisions during critical patient outcomes (discharge or crisis). The lower entropy of 0.69 further suggests that the geometric penalty helps the policy converge to a more confident and consistent treatment regime.

As for the safety of the policies we observe standard IQL achieves the lower deviation from clinical behaviour as can be seen from Table 3. This extreme adherence limits the ability to improve upon the behavioural clinician policy. 

Geo-IQL exhibits marginally higher deviation yet successfully remains well within safe bounds of extreme action agreement with 97.86\%.

Crucially when comparing Q-Improvement, $\Delta Q$, we see that Geo-IQL is able to effectively break out the local minima with the slight deviations in actions that typically trap IQL. This further supported by the lower Policy-Clinician Agreement rates. The geometry based penalty allows the agent to safely explore high-value actions that differ from the clinician, whereas IQL is \textit{implicitly} being regularised to the behavioural policy by design.

\section{Conclusion}
This study addresses the critical challenge of Q-Value overestimation in Offline RL by introducing a geometric, data-centric penalty mechanism. By decoupling the uncertainty estimation from the policy training loop, we achieve a method that is both computationally efficient and outperforms current top performing algorithms.

Our experimental analysis yield three primary conclusions:
\begin{enumerate}
    \item Stability on Fractured Manifolds: In hyper-sensitive dynamics of \texttt{Hopper-Medium-replay-v2}, where standard algorithms wildly oscillate between sustained success and instant failure, Geo-IQL stabilises learning, while outperforming computationally expensive baselines with the speed of IQL.
    \item Strategic optimisation in Healthcare: On the MIMIC-III Sepsis dataset we demonstrate that standard IQL behaves as over-regulariser, achieving high average agreement but failing to improve patient outcomes. Geo-IQL successfully breaks this plateau by selective penalising unsupported actions, allowing the policy to deviate strategically in intermediate steps, optimising vasopressor dosages while converging more reliably to successful outcomes, the higher terminal agreement.
    \item Efficiency: Unlike iterative regularisation methods like CQL that require sampling every training steps our precomputed lookup approach maintains a $\mathcal{O}(1)$ runtime overhead, making it scalable for large high-dimensional real-world datasets.
\end{enumerate}

In summary, Geometric Pessimism guides the policy to safe regions on the state-action space with density-adaptive penalisation, where the agent can confidently improve upon baselines. This work further proves the theoretical backing and bridges the gap between the theoretical conservatism and practical, safe deployment of RL in critical systems like robotics and healthcare.

\clearpage
\bibliographystyle{apalike}
\bibliography{references}

\clearpage
\appendix
\section{Theoretical Proof}

\subsection{Setup \& Definitions}

\begin{definition}[State-Action Space]
$\mathcal{S} \times \mathcal{A}$
\end{definition}

\begin{definition}[Offline Dataset]
$\mathcal{D} = \{(s_i, a_i, r_i, s'_i)\}_{i=1}^N$
\end{definition}

\begin{definition}[True Q-function]
$Q^*: \mathcal{S} \times \mathcal{A} \rightarrow \mathbb{R}$
\end{definition}

\begin{definition}[Learned Q-function]
$\widehat{Q}: \mathcal{S} \times \mathcal{A} \rightarrow \mathbb{R}$
\end{definition}

\begin{definition}[Distance to Dataset]
\[
d((s,a), \mathcal{D}) := \min_{(s',a') \in \mathcal{D}} \|(s,a)-(s',a')\|_2
\]
\end{definition}

\begin{definition}[Geo-IQL Estimate]
\[
\tilde{Q} := \widehat{Q} - \lambda \cdot d((s,a), \mathcal{D})
\]
\end{definition}

\subsection{Assumptions}

\begin{enumerate}
    \item $Q^*$ is $L$-Lipschitz continuous:
    \begin{equation*}
    |Q^*(s_1,a_1) - Q^*(s_2,a_2)| \leq L^* \cdot \|(s_1,a_1) - (s_2,a_2)\|
    \end{equation*}
    for all $(s_1,a_1), (s_2,a_2) \in \mathcal{S} \times \mathcal{A}$.

    \item $\widehat{Q}$ is $L$-Lipschitz continuous:
    \begin{equation*}
    |\widehat{Q}(s_1,a_1) - \widehat{Q}(s_2,a_2)| \leq \widehat{L} \cdot \|(s_1,a_1) - (s_2,a_2)\|
    \end{equation*}
    for all $(s_1,a_1), (s_2,a_2) \in \mathcal{S} \times \mathcal{A}$.

    \item $\widehat{Q}$ approximates $Q^*$ well on $\mathcal{D}$:
    \begin{equation*}
    |\widehat{Q}(s,a) - Q^*(s,a)| \leq \epsilon
    \end{equation*}
    for all $(s,a) \in \mathcal{D}$.
\end{enumerate}

Assumptions 1 \& 2 are reasonable given their neural network structure and ReLU \cite{virmaux2018lipschitz}.
Assumption 3 is also reasonable given we are training on that data itself.

\subsection{Theorem Statement}
\begin{theorem}
\label{thm:main}
Under Assumptions 1 and 2, for any $(s,a) \in \mathcal{S} \times \mathcal{A}$, if the penalty parameter satisfies
\[
\lambda \geq \widehat{L} + L^* + \frac{\epsilon}{d_{\min}},
\]
where 
\[
d_{\min} = \min_{(s',a') \notin \mathcal{D}} \|(s,a)-(s',a')\|_2,
\]
then \[\tilde{Q}(s,a) \leq Q^*(s,a)\]
\end{theorem}

\subsection{Proof}

\begin{figure}[t]
\begin{center}
\centerline{\includegraphics[width=\columnwidth]{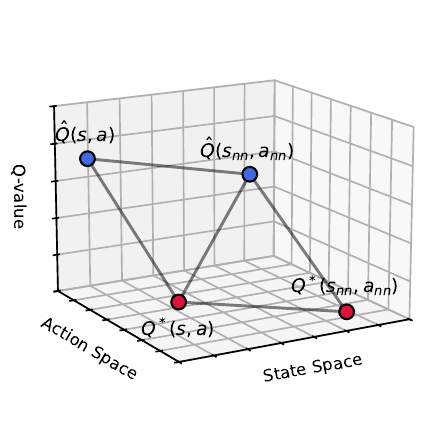}}
\caption{Triangle Inequality}
\end{center}
\vskip -0.2in
\end{figure}

\subsubsection{Step 1: Define Nearest Neighbour}

Let $(s_{nn}, a_{nn}) := \arg \min_{(s',a') \in \mathcal{D}} \|(s,a)-(s',a')\|_2$.

Let $d := d((s,a), \mathcal{D}) = \|(s,a)-(s_{nn},a_{nn})\|_2$.

\subsubsection{Step 2: Decomposition via Triangle Inequality}
See Fig 8. for visual intuition.
\begin{align*}
|\widehat{Q}(s,a) - Q^*(s,a)| &\leq |\widehat{Q}(s,a) - \widehat{Q}(s_{nn},a_{nn})| \\
&\quad + |\widehat{Q}(s_{nn},a_{nn}) - Q^*(s_{nn},a_{nn})| \\
&\quad + |Q^*(s_{nn},a_{nn}) - Q^*(s,a)|
\end{align*}

\subsubsection{Step 3: Bounding Each Term}

\textbf{Term 1:} By Assumption 2 ($\widehat{Q}$ is $\widehat{L}$-Lipschitz):
\[
|\widehat{Q}(s,a) - \widehat{Q}(s_{nn},a_{nn})| \leq \widehat{L} \cdot d
\]

\textbf{Term 2:} By Assumption 3, since $(s_{nn}, a_{nn}) \in \mathcal{D}$:
\[
|\widehat{Q}(s_{nn},a_{nn}) - Q^*(s_{nn},a_{nn})| \leq \epsilon
\]

\textbf{Term 3:} By Assumption 1 ($Q^*$ is $L^*$-Lipschitz):
\[
|Q^*(s_{nn},a_{nn}) - Q^*(s,a)| \leq L^* \cdot d
\]

\textbf{Combined Bound:}
\[
|\widehat{Q}(s,a) - Q^*(s,a)| \leq (\widehat{L} + L^*) \cdot d + \epsilon
\]

\subsubsection{Step 4: Extracting One-Sided Bound}

From the absolute value bound:
\[
\widehat{Q}(s,a) - Q^*(s,a) \leq (\widehat{L} + L^*) \cdot d + \epsilon
\]

Rearranging:
\[
\widehat{Q}(s,a) \leq Q^*(s,a) + (\widehat{L} + L^*) \cdot d + \epsilon
\]

\subsubsection{Step 5: Applying the Geo-IQL Penalty}

\begin{align*}
\tilde{Q}(s,a) &= \widehat{Q}(s,a) - \lambda \cdot d \\[6pt]
&\leq Q^*(s,a) + (\widehat{L} + L^*) \cdot d + \epsilon - \lambda \cdot d \\[6pt]
&= Q^*(s,a) + \epsilon - (\lambda - \widehat{L} - L^*) \cdot d
\end{align*}

\subsubsection{Step 6: Establishing Pessimism Condition}

For $\tilde{Q}(s,a) \leq Q^*(s,a)$, we require:
\[
\epsilon - (\lambda - \widehat{L} - L^*) \cdot d \leq 0
\]

Solving for $\lambda$:
\[
\lambda \geq \widehat{L} + L^* + \frac{\epsilon}{d}
\]

Since $d \geq d_{\min}$ for all out-of-distribution points:
\[
\boxed{\lambda \geq \widehat{L} + L^* + \frac{\epsilon}{d_{\min}}}
\]

\subsubsection{Conclusion}

Under the stated condition on $\lambda$, we have:
\[
\tilde{Q}(s,a) \leq Q^*(s,a) \quad \forall (s,a) \in \mathcal{S} \times \mathcal{A}
\]

This establishes that the Geo-IQL estimate is a guaranteed pessimistic lower bound on the true optimal Q-function. \hfill $\blacksquare$

\section{Codebase}
Github Repository : 
\url{https://github.com/swanjax/SSEF_RO053}

The environment requirements for D4RL and MuJoCO require certain versions of dependencies. We direct the reader to the CORL implementation referenced.

\end{document}